\documentclass{article}

\PassOptionsToPackage{numbers, compress}{natbib}

    \usepackage[preprint]{neurips_2025}

\usepackage[utf8]{inputenc} 
\usepackage[T1]{fontenc}    
\usepackage{hyperref}       
\usepackage{url}            
\usepackage{booktabs}       
\usepackage{amsfonts}       
\usepackage{nicefrac}       
\usepackage{microtype}      
\usepackage{xcolor}         

\title{Priors Matter: Addressing Misspecification in Bayesian Deep Q-Learning}

\usepackage{amsmath}
\usepackage{amssymb}
\usepackage{mathtools}
\usepackage{amsthm}

\author{%
  Pascal R. van der Vaart \\
  Delft University of Technology\\
   2628 CD Delft, The Netherlands\\
  \texttt{p.r.vandervaart-1@tudelft.nl} \\
  \And
  Neil Yorke-Smith \\
  Delft University of Technology\\
  2628 CD Delft, The Netherlands\\
  \texttt{n.yorke-smith@tudelft.nl} \\
  \AND
  Matthijs T. J. Spaan\\
  Delft University of Technology\\
  2628 CD Delft, The Netherlands\\
  \texttt{m.t.j.spaan@tudelft.nl} \\
}
\begin{document}

\maketitle

\begin{abstract}
    Uncertainty quantification in reinforcement learning can greatly improve exploration and robustness. Approximate Bayesian approaches have recently been popularized to quantify uncertainty in model-free algorithms. However, so far the focus has been on improving the accuracy of the posterior approximation, instead of studying the accuracy of the prior and likelihood assumptions underlying the posterior. In this work, we demonstrate that there is a cold posterior effect in Bayesian deep Q-learning, where contrary to theory, performance increases when reducing the temperature of the posterior. To identify and overcome likely causes, we challenge common assumptions made on the likelihood and priors in Bayesian model-free algorithms. We empirically study prior distributions and show through statistical tests that the common Gaussian likelihood assumption is frequently violated. We argue that developing more suitable likelihoods and priors should be a key focus in future Bayesian reinforcement learning research and we offer simple, implementable solutions for better priors in deep Q-learning that lead to more performant Bayesian algorithms.
\end{abstract}

\section{Introduction}
\label{sec:intro}

Reinforcement learning (RL) algorithms have many potential applications, but the exploration--exploitation trade off remains an open problem. Especially when real or simulated experiences are expensive, it is essential that RL agents can efficiently explore the environment to increase sample efficiency. Many exploration methods rely on the quantification of uncertainty, by assigning novelty bonuses \citep{ostrovski2017count, bellemare2016unifying, burda2018exploration} or through sampling approaches such as Thompson sampling \citep{osband2016deep, osband2018randomized, o2018uncertainty, fortunato2017noisy, schmitt2023exploration,  azizzadenesheli2018efficient, dwaracherla2021langevin}. However, quantification of uncertainty for deep RL remains a challenging problem. 

One uncertainty quantification method is through \emph{Bayesian inference}, where an agent learns how likely certain models or values are, given a prior and the data it has observed. In theory, Bayesian algorithms have strong theoretical guarantees in well-defined settings. A well-known result in statistical learning theory is that Bayesian algorithms achieve optimal average loss with the correct prior and likelihood \citep{komaki1996}.  Further, in bandit settings and the model-based RL, Thompson sampling is proven to have strong regret bounds \citep{agrawal2012analysis,osband2013more}.

However, in the benchmarks currently used in deep reinforcement learning the performance of Bayesian approaches depends heavily on the environment \citep{ishfaq2023provable, bayesian2024vaart}, and they are sometimes outclassed by straightforward ensembles of maximum likelihood estimators, such as BootDQN \citep{osband2016deep, osband2018randomized}. The difference between practice and theory in reinforcement learning could be due to a variety of challenges that deep reinforcement learning imposes. This discrepancy in performance is not unique to reinforcement learning, however. 

In deep supervised learning, \citet{wenzel2020good} identified a \emph{cold posterior effect}, where performance increases as the temperature of the target posterior is decreased. This clashes with the statistical learning point of view that states that the Bayesian posterior should provide optimal performance \citep{aitchison1975goodness,komaki1996,zellner1988optimal}. Markov Chain Monte Carlo (MCMC) methods tailored to deep learning have recently been developed and shown to have sufficient performance \citep{wenzel2020good}, implying that the cold posterior effect is not due to poor approximation methods. 

Another possible cause is misspecification of the model, i.e. the assumed likelihood and prior. Due to the complexity of neural networks, it is near impossible to translate a real-world prior back into a prior over network parameters, and typically simple Gaussian priors are picked. Indeed, \citet{fortuin2022bayesian} find in supervised learning that improving the choice of prior can reduce the cold posterior effect.

In deep reinforcement learning, misspecification of priors has been understudied. Recent methods have used Gaussian priors \citep{dwaracherla2021langevin, schmitt2023exploration, ishfaq2023provable, bayesian2024vaart}, and the \textbf{choice of prior} is left as an afterthought and has not gained much attention. In our work however, we find that Gaussian priors \emph{are} misspecified, and that improving the prior leads to more performant Bayesian deep RL algorithms.

Furthermore, another open issue is the \textbf{choice of likelihood} in RL. Supervised learning learns from ground-truth labels, often providing obvious likelihood assumptions. For example, a categorical distribution for a multi-class classification task. On the other hand, reinforcement learning methods often learn by minimizing the difference of the current value estimate and a self-supervised (bootstrapped) value estimate of the next state, known as the temporal difference error. A likelihood on these temporal difference errors is difficult to choose correctly, as their distribution depends on the reward function, transition function and the policy itself.

Previous work in Bayesian model-free RL has typically assumed that temporal difference errors follow a normal distribution \citep{Dearden, ishfaq2023provable, bayesian2024vaart, schmitt2023exploration, dwaracherla2021langevin, azizzadenesheli2018efficient}, likely due to ease of inference, or as a standard Bayesian extension to the squared temporal difference error that maximum likelihood approaches would use. However, as we demonstrate, this is not a realistic assumption in many benchmark tasks. In non-Bayesian RL  there has been increasing interest in other losses, specifically the logistic loss~\citep{bas2021logistic, lv2024modeling}, after observing that the distribution of TD errors are closer to a logistic distribution than a normal distribution.

In this work, we establish the \textbf{existence of the cold posterior effect in Deep Q-Learning (DQN)}, and investigate potential causes. We extensively test prior and likelihood assumptions on a wide range of benchmark tasks.
We find experimentally that despite widespread adoption, Gaussian priors are not a good fit in RL methods. To remedy this, we introduce Laplace priors to RL, and show that they are a better fit and provide higher performance at very low computational and implementation cost. Furthermore, we demonstrate the feasibility of meta-learning a prior on different tasks that generalizes to the test task, achieving higher performance than both Gaussian and Laplace priors.  Finally, we demonstrate through statistical tests that the distribution of TD errors is neither a normal nor a logistic distribution, and discuss the complications of choosing better likelihoods.
Our work establishes the development of more performant priors and likelihoods as a viable future research direction to improve the performance of deep reinforcement learning algorithms.

\section{Background}
\label{seg:bg}

\subsection{Reinforcement Learning}
We consider the standard discounted infinite horizon MDP \citep{sutton2018reinforcement}, which is a tuple $(S, A, R, T, \gamma)$ consisting of a state space $S$, action space $A$, deterministic reward function $R$ and transition function $T$ and discount factor $0 < \gamma < 1$.

At each time step $t$, the agent receives the current state $s_t \sim T(s_{t-1}, a_{t-1})$, chooses an action $a_t \sim \pi(s_t)$ from its policy $\pi$, and receives reward $r_t = R(s_t, a_t)$. The goal is to find a policy $\pi$ that maximizes the expected cumulative discounted reward $\mathbb{E}[J(\pi)] = \mathbb{E}\left[\sum_{t=0}^\infty \gamma^t r_t\right].$

Of central importance is the Q-value function $Q^\pi(s, a) = R(s, a) + \sum_{t=1}^\infty \gamma^t r_t$, which maps a state-action to the future expected cumulative discounted reward after executing action $a$ in state $s$, and executing a given policy $\pi$ afterwards. The Q-function satisfies a recursive relationship
\begin{align}
\begin{split}
            Q^\pi(s, a) = R(s, a) +
             \gamma \mathbb{E}\left[Q^\pi(s', a') | s' \sim T(s, a), a' \sim \pi(s') \right],
\end{split}
\end{align} 

\subsection{Bayesian Value Learning} \label{sec:bayesian-value-learning}
With a parameterized model $Q_\theta$, Bayesian algorithms aim to infer the posterior
$$p(\theta | \mathcal{D}) = \frac{p(\mathcal{D} | \theta)p(\theta)}{\int p(\mathcal{D} | \theta)p(\theta)d(\theta)},$$ 
with $p(\mathcal{D} | \theta)$ representing the likelihood, $p(\theta)$ the prior, and $\mathcal{D}$ the observed data. The posterior, $p(\theta | \mathcal{D})$, describes the plausibility of parameter values, making it a natural way to express uncertainty.

To equip an RL agent with the ability to quantify uncertainty over values, we can construct a posterior over the parameters of a Q-function as 
$p(\theta | \mathcal{D}) \propto p(\mathcal{D} | \theta)p(\theta)$.
Given that the squared error loss is proportional to the log-probability of a normal distribution, an evident choice for the likelihood in a Bayesian formulation of value-based algorithms is
\begin{equation}\label{eq:tddensity}
    p(\mathcal{D} | \theta) = \exp\left(-\sum_{\ \ {(s, a, r, s') \in \mathcal{D}}} \left[Q_\theta(s, a) - r - 
\gamma G(\theta, s')\right]^2 \right),
\end{equation}
where $G(\theta, s')$ is some estimator for the (optimal) return at $s'$, possibly bootstrapped from our model $\theta$.
This corresponds to the assumption that temporal difference (TD) errors follow a normal distribution:
\begin{equation}\label{eq:tdnormal}
    \mathit{TD}(\theta, (s, a, r, s')) \sim \mathcal{N}(0, \sigma).
\end{equation}
Although this assumption may not be valid for every MDP, it is a convenient design decision in deep RL, and it is not surprising that various previous works have employed it \citep{osband2018randomized, schmitt2023exploration, dwaracherla2021langevin, azizzadenesheli2018efficient, ishfaq2023provable}. The prior is usually also chosen to be a normal distribution, which corresponds to using $\ell_2$ regularisation in maximum likelihood estimation.
The likelihood and prior, together with an inference method, define which posterior a Bayesian RL algorithm ends up with.

\subsection{Using Posterior Distributions for Exploration}\label{sec:posterior_using}
Equipped with a posterior distribution signifying how likely a given model is, an agent can explore through a variety or techniques. Most common are optimism-based algorithms, where an attempt is made to upper bound the value for each action with some confidence level, and then explore by taking the action with the highest current upper bound. This method is widely studied in both bandit settings~\citep{lai1985ucb} and RL~\citep{auer2008near}, achieving good theoretical performance. Another approach is Thompson sampling (TS), where the posterior is sampled and the agent acts greedily with respect to the sample for one action or episode, also achieving good theoretical performance in both bandits~\citep{agrawal2012analysis} and RL~\citep{osband2013more}.

Crucially, the approximated posterior needs to be close to the true posterior for these methods to work well. For optimism, an incorrect posterior will lead to incorrect bounds. Bounds that are too loose will result in less efficiency, causing an agent to perhaps execute an action too many times before lowering the bound to a suitable level. On the other hand, bounds that are too tight can be more catastrophic, causing an agent to never execute an action even though it might be the optimal action. Thompson sampling suffers from the same issues, where a posterior that does not contract fast enough leads to over-exploration, and contracting too fast leads to under-exploration. Thompson sampling with misspecified prior distributions has been studied in a bandit setting \citep{simchowitz2021bayesian}, suffering a penalty based on the difference between the chosen prior and the true prior.

\subsection{Cold Posterior Effect}
The cold posterior effect refers to the phenomenon where Bayesian neural networks (BNNs) sometimes achieve superior predictive performance when their posterior distributions are artificially `cooled' by exponentiating the density by a temperature parameter $T < 1$:
\begin{equation}
    p(\theta | \mathcal{D})^\frac{1}{T} \propto (p(\mathcal{D} | \theta)p(\theta))^\frac{1}{T}.
\end{equation}
As $T$ decreases, the inverse temperature $\frac{1}{T}$ increases, effectively sharpening the posterior and therefore deliberately underestimating uncertainty. While in theory $T=1$ is expected to be optimal \citep{aitchison1975goodness, komaki1996,zellner1988optimal}, \citet{wenzel2020good} find that decreasing the temperature increases performance in deep supervised classification settings. They attribute this to misspecification of the prior distributions. On the other hand, \citet{aitchison2021a} concludes that likelihoods are also misspecified in supervised learning, and \citet{izmailov2021bayesian} identify data augmentation as a cause. Recently, \citet{mclatchie2024predictive} theoretically proved under several conditions, including a large sample size, that tempering the posterior does not influence the predictive performance. Reinforcement learning with Bayesian neural networks poses an interesting point of view, since the sample size is initially small, and agents performance relies both on uncertainty quantification and predictive performance.

 \begin{figure}[t]
     \centering
     \includegraphics[width=1.0\textwidth]{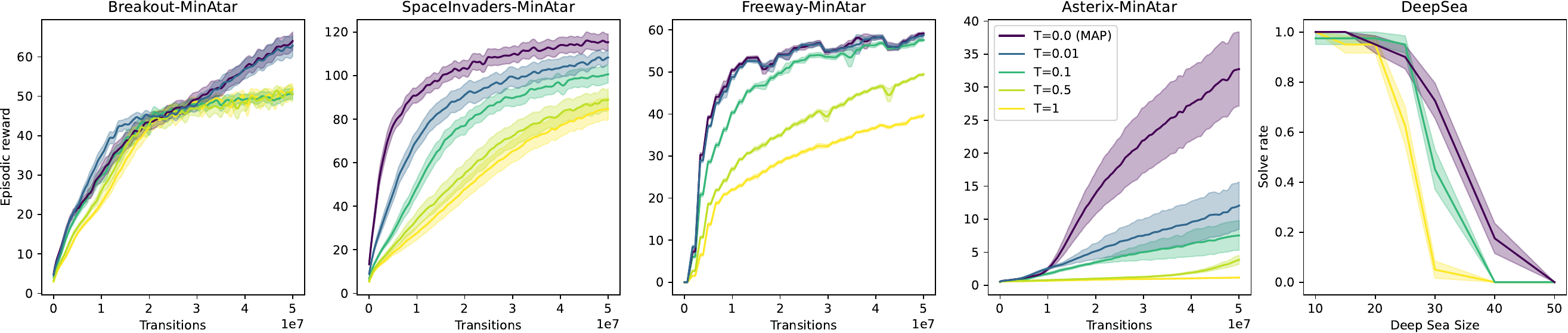}
     \caption{\textbf{Left four plots:} Performance of Bayesian Q-learning for five different temperatures on MinAtar. The line is the mean of 50 seeds, with shaded area displaying one standard error of the mean. There is a clear correlation between lower temperature and high performance.\\
     \textbf{Right:} Solve rate of Bayesian DQN on Deep Sea at several sizes with a Gaussian prior and likelihood after 200K episodes at several temperatures. The solve rate is computed over 40 independent seeds. The shaded area displays one standard error.}
     \label{fig:minatar-cold-posterior}
 \end{figure}

\section{The Cold Posterior Effect in DQN}\label{sec:coldposterior-dqn}

Here, we demonstrate that Bayesian DQN methods also suffer from a cold posterior effect. In contrast to supervised learning, RL presents a unique situation, as underestimating uncertainty can lead to poor exploration during training, potentially leading to a large drop in performance. Furthermore, reinforcement learning methods are more susceptible to misspecification of the likelihood. The Gaussian assumption on temporal difference errors is likely incorrect on many benchmarks. For example, a deterministic environment with a deterministic optimal policy will lead to deterministic temporal difference errors. Finally, while data augmentation has been used in RL \citep{laskin2020reinforcement}, it is less common and we forego any data augmentation techniques in this work.

While we should expect better performance from our Bayesian algorithm at $T=1$, we see very clearly from our experiments that reducing the target temperature improves performance. For this experiment we ran our Bayesian Q-learning algorithm, introduced in Section~\ref{sec:method} at $T \in \{0, 0.01, 0.1, 0.5, 1\}$. We tuned the hyperparameters to perform well on Breakout-Minatar at $T=1$, and then test with the same hyperparameters for $T=0$ and $T=0.1$ in all MinAtar environments. The hyperparameters are optimized through Bayesian search, with details in Appendix~\ref{sec:hyperparameters}.

Figure~\ref{fig:minatar-cold-posterior} clearly shows
that reducing the posterior temperature improves performance, even on the task for which the hyperparameters were tuned. Setting the temperature to $T=0$ essentially reduces the sampler to maximum a posteriori (MAP) estimation, equivalent to minimizing the squared TD-loss with regularization. Furthermore, Figure~\ref{fig:minatar-cold-posterior} shows that even on Deep Sea \citep{osband2020Behaviour}, an exploration task, an ensemble of MAP estimates has an advantage over the posterior. In the next sections we highlight two potential problems that cause this effect: misspecified priors and misspecified likelihoods.

\begin{figure}
    \centering
    \includegraphics[width=1.0\linewidth]{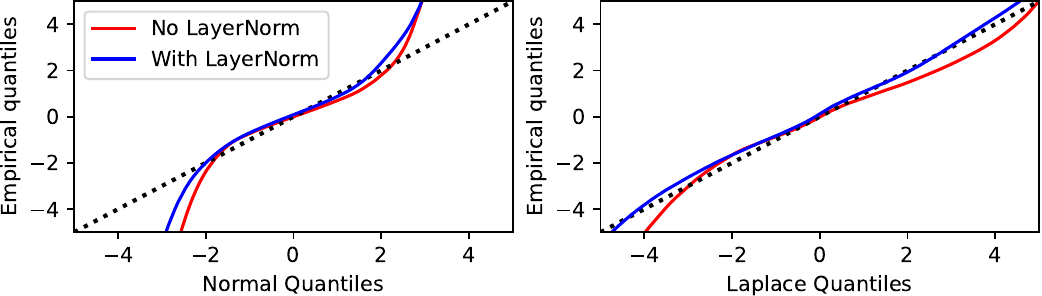}
    \caption{Q-Q plots with respect to a normal {\bf (left)} or Laplace {\bf (right)} of hidden layer weights of a Q-network after training with and without LayerNorm, aggregated over all environments. We can see that Laplace distributions are a much closer fit than normal distributions, which have flatter tails.}
    \label{fig:priors_all}
\end{figure}

\section{Are Priors Misspecified?}
The Bernstein von Mises' theorem states that as more data is collected, the influence of the prior will eventually fade under regularity conditions. Nonetheless, a misspecified prior leads to sub-optimal regret in bandit settings \citep{simchowitz2021bayesian}. 
While \citet{fortuin2022bayesian} has previously concluded that priors in deep supervised learning are misspecified, this has not yet been studied in a reinforcement learning setting. We hypothesize that the typically used Gaussian priors in reinforcement learning \citep{dwaracherla2021langevin, schmitt2023exploration, bayesian2024vaart} are in fact also misspecified and a likely cause for the discrepancy in performance of Bayesian DQN.
\subsection{Prior Misspecification}
To test our hypothesis, we train multiple Q-learning agents and inspect the distributions of their parameters after training. Ideally, aggregating all parameters over multiple benchmark tasks would yield a distribution close to the assumed prior. Figure~\ref{fig:priors_all} shows the empirical distribution over the parameters of the second layer in a 3-layer fully connected Q-network, aggregating over 18 environments with discrete actions in the Gymnax benchmark \citep{gymnax2022github}.

We can see in the Q-Q plot that the empirical distribution over parameters is more heavy tailed than a normal distribution, signifying that a Gaussian prior might neglect certain parameter configurations that are realistic outcomes in practice. In other words, a Gaussian prior in a Bayesian DQN algorithm can actively hinder the agent from learning the true optimal values.  \citet{fortuin2022bayesian} find a similar result in classification tasks, signifying that priors for supervised learning tasks might generalize to reinforcement learning.  We also plot against the quantiles of a Laplace distribution, and observe a much closer fit.

\subsection{Improving the Prior}

\paragraph{Using Layer Normalization}
Layer normalization \citep{ba2016layer} is a popular normalization method that normalizes the activations in a neural network, and then explicitly rescales them before applying the activation function. Recently, \citet{gallici2024simplifying} demonstrated that layer normalization has theoretical stabilizing properties in Deep Q-learning, with visible practical advantages. Using layer normalization also provides advantages when picking a prior, by causing the outputs of a layer to be invariant under scaling of the weight matrix. As a result, the choice of prior for the weight matrix can be simplified by eliminating the need to set the scale. However, layer normalization introduces its own parameters that require a prior, and normal distributions still have an improper shape according to Figure~\ref{fig:priors_all}.

\paragraph{Using a Laplace distribution}
As shown in Figure~\ref{fig:priors_all}, the Laplace distribution is a much better fit to the parameter distribution that we found empirically, especially in combination with layer normalization. Therefore, simply replacing the Gaussian prior with a Laplace prior can be expected to lead to improved results. We test this hypothesis in Section~\ref{sec:experiments}.

\paragraph{Meta-learning a Prior}
While a Laplace distribution is a closer fit to the empirical distribution on the weights of the hidden layer, it is likely to be affected by the chosen activation function, position of the layer in the network, and architecture or function of the layer (e.g., convolutional, attention, layer norm). Therefore, we develop a more flexible approach for specifying priors for specific architectures. We fit a small scalar normalizing flow \citep{rezende2015variational} to the empirical distribution individually for each layer's parameters. The weights of a single layer are then assumed to be drawn i.i.d.\@ from each layer's corresponding normalizing flow. The normalizing flow is a scalar distribution and only consists of one spline with two knots and two affine layers, causing minimal extra computational cost. We test the resulting priors in Section~\ref{sec:experiments}.
 \begin{figure}[tb]
     \centering
     \includegraphics[width=1.0\linewidth]{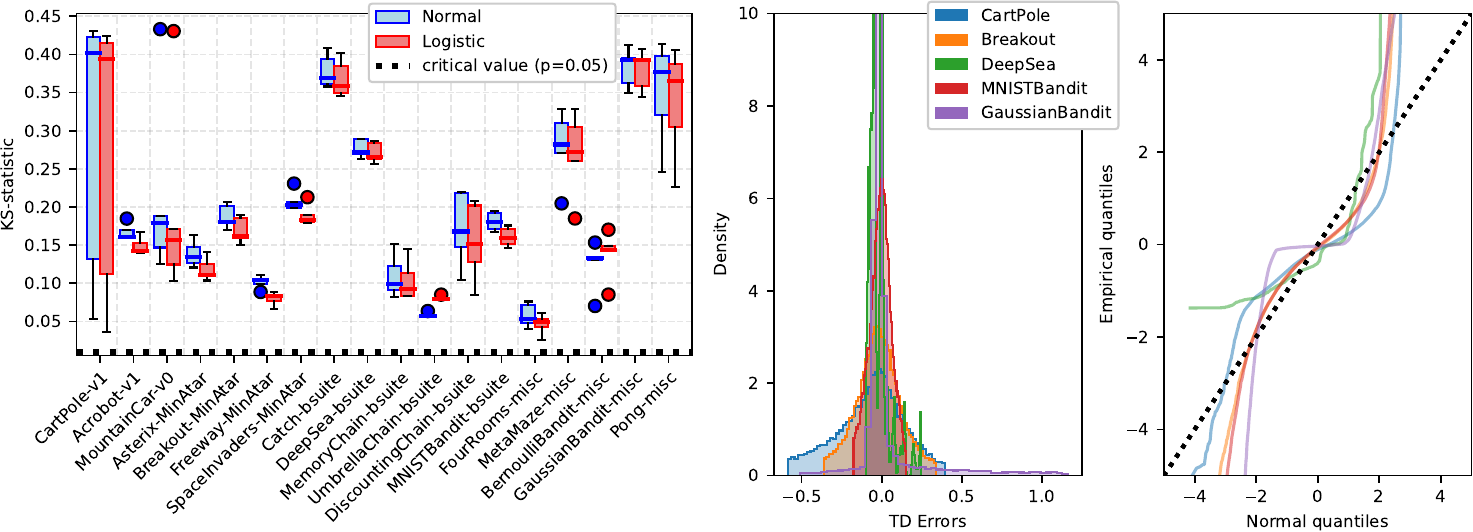}
     \caption{\textbf{Left:} Boxplots depicting 10 repititions of Kolmogorov-Smirnov statistics tested against both normal and logistic distributions for TD errors of Q-learning on 19 Gymnax environments. The horizontal dashed line indicates the critical value for $p=0.05$, which is obtained through simulation for both normal and logistic distributions. \textbf{Middle, Right}: Histograms and Q-Q plots of empirically observed temporal difference errors for Q-learning agents in 5 environments from Gymnax. The Q-Q plots are rescaled to mean $0$ and standard deviation $1$.}
     \label{fig:likelihoods_compare}
 \end{figure}

 \begin{figure}[tb]
    \centering
    \includegraphics[width=0.90\linewidth]{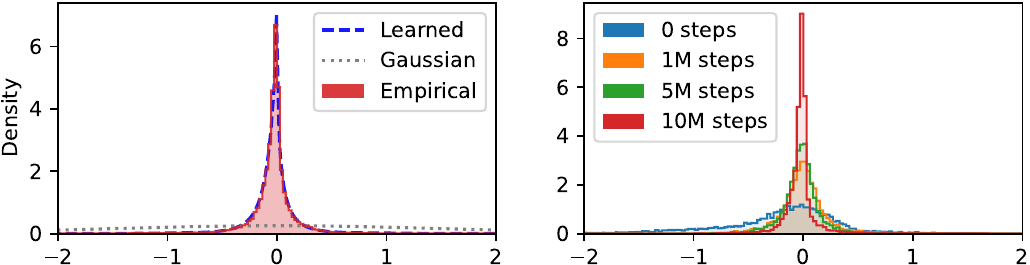}
    \caption{\textbf{Left:} Empirical TD errors on Breakout-MinAtar observed by a DQN agent, together with the learned likelihood model and for comparison a normal distribution with standard deviation 1.6, which is tuned for performance of our Bayesian DQN agent, but clearly does not represent the true empirical distribution well.
    \textbf{Right:} Empirical distributions of temporal difference errors after zero, one million, five million and ten million steps.}
    \label{fig:likelihood-breakout}
\end{figure}

\section{Are Likelihoods Misspecified?}
Wrong likelihoods lead to incorrect posterior contraction and incorrect credible sets, even in the limit~\citep{kleijn2012bernstein}. This means that TS-style algorithms are no longer sampling actions proportional to their chance of optimality, and UCB-style algorithms are operating with incorrect bounds. Furthermore, even with infinite data, the maximum likelihood estimator under a misspecified likelihood is likely to be different from the true maximum likelihood estimator, meaning that the posterior distribution is also mispositioned.

\subsection{Gaussian Likelihoods}
The common choice of a Gaussian likelihood can be attributed to two reasons. First, typical DQN algorithms minimize the squared TD error, which is equivalent to maximizing the log-density of a normal distribution:
\begin{align}
\arg \min_\theta \sum_{\ \ \mathclap{(s, a, r, s') \in \mathcal{D}}} \left[Q_\theta(s, a) - r - 
\gamma G(\theta, s')\right]^2 &= \arg \max_\theta \sum_{\mathcal{D}} \frac{-1}{2\sigma_{\mathit{TD}}^2}\left[Q_\theta(s, a) - r - \gamma G(\theta, s')\right]^2 \\ 
&= \arg \max_\theta \log p(\mathcal{D} | \theta).
\end{align} 

Taking the point of view that DQN is a frequentist maximum likelihood approach that optimizes a Gaussian likelihood $p(\mathcal{D} | \theta)$, 
a natural practical Bayesian extension is then to pick some prior $p(\theta)$ and infer a posterior $\log p(\theta |\mathcal{D}) \propto \log p(\mathcal{D} | \theta) + \log p(\theta)$.

A more theoretical motivation for a normally distributed likelihood can be traced back to \citet{Dearden}. The Q-values are large sums of discounted rewards $Q^\pi = \sum_{i=1}^\infty \gamma^i r_i$, so by the central limit theorem \citep{vaart2000asymptotic} their distribution should resemble a normal distribution if the MDP is ergodic under the optimal policy. However, this argument unfortunately does not extend to our current situation, as the typical benchmarks are not ergodic and often even episodic. Furthermore, many tasks have sparse rewards where many $r_i$ are equal to $0$. Also, the $r_i$ are not actually independent variables, since by the Markov property they are only independent when conditioning on the state $s_i$. 

Finally, even if $Q(s, a)$ and $Q(s', a)$ for consecutive states $s, s'$ are both normally distributed, there is no guarantee that $Q(s, a) - r - \gamma Q(s', a)$ is normally distributed because $Q(s, a)$ and $Q(s', a)$ are not independent random variables. In fact, they sum over the same future rewards $r_i$. \citet{Dearden} do make the assumption that these are independent in their Assumption 4, but also highlight that this assumption is generally false. 

Recently, the logistic loss has gained some popularity for DQN \citep{bas2021logistic, lv2024modeling}, where \citet{lv2024modeling} find that the logistic distribution is closer to the true error distribution than a normal distribution. In our work however, we empirically observe that neither the normal nor logistic distribution is a statistically correct choice.

\paragraph{Empirical Validation}
To investigate whether our typically assumed likelihoods are valid, we train multiple Q-learning agents until convergence, and track the temporal difference errors observed at the end of training. We statistically test whether these errors come from a normal or logistic distribution using a Kolmogorov-Smirnov (KS) test \citep{daniel1990applied}, while simultaneously estimating the parameters of the test distribution. This means that we are testing whether the TD errors come from any normal or logistic distribution. We highlight that this is a luxury that RL agents typically do not have: the likelihood scale is usually a fixed hyperparameter and has to be guessed (or tuned) correctly in advance. We refer to Appendix~\ref{sec:ks-test} for details on the specific KS test that we used.

The left plot in Figure~\ref{fig:likelihoods_compare} shows that on every environment, the null hypothesis can be rejected, meaning that the TD errors follow distributions that are significantly different from both a normal and a logistic distribution. Furthermore, perhaps more troubling are the right plots in Figure \ref{fig:likelihoods_compare}, which shows that each environment in our benchmark set induces vastly different distributions for the TD errors. The Q-Q plot on the right shows that the distributions vary significantly even when correcting for the mean and standard deviation individually per environment. This means that knowing the parameters of the likelihood ahead of time, which are usually hyper parameters of an algorithm, does not guarantee a good fit. The different shapes mean it could be very difficult to construct a single likelihood that can be expected to generalize over many tasks.

\subsection{Improving the Likelihood}
It is tempting to improve the choice of likelihood by investigating empirical TD error distributions, however we have already seen in Figure~\ref{fig:likelihoods_compare} that the TD error distributions differ greatly per environment. Fitting one likelihood per environment is unfeasible in practice, as we are interested in the distribution of TD errors under the optimal policy. This means that fitting an empirical likelihood requires a pre-trained agent for each environment -- defeating the purpose. 

Nonetheless, to study whether having oracle access to this distribution would aid the agent in practice, we fit a distribution to the empirical data of each MinAtar environment, and test Bayesian DQN with the assumed likelihood. As a main potential problem, we highlight that the likelihood plays both the role of uncertainty quantification and that of a loss function. For example, Figure~\ref{fig:likelihood-breakout} shows the likelihood for Breakout-MinAtar, which has a much sharper peak than a normal distribution. It is likely that this will cause an ill-conditioned loss landscape for gradient based optimization. Another problem when choosing a likelihood is that the distribution of actually observed temporal difference errors changes during training as shown in Figure~\ref{fig:likelihood-breakout}.  The wider distribution at the start can cause very large gradients under the sharply peaked likelihood in both the $T=0$ and $T=1$ agent. We thus hypothesize that agents with correct likelihood distributions will not necessarily do well from a performance point of view.

\section{Empirical Study}
\label{sec:experiments}
We introduce our Bayesian DQN implementation, and empirically test our proposed solutions to the problems with priors and likelihoods in Bayesian DQN. 
\subsection{Algorithm Tested}\label{sec:method}
While several deep Bayesian Q-learning methods exist \citep{dwaracherla2021langevin, ishfaq2023provable, bayesian2024vaart, azizzadenesheli2018efficient, schmitt2023exploration}, we pose that these are all special cases of the outline in Section~\ref{sec:bayesian-value-learning}. That is, they pick a return estimator $G(\theta, s')$ and a likelihood on the temporal difference error $Q_\theta(s, a) - r - \gamma G(\theta, s')$, and then use a specific inference method to approximate the posterior distribution. For example, the previously mentioned papers all take $G(\theta, s') = \max_a Q_\theta(a, s')$ and assume a normal distribution as likelihood. \citet{schmitt2023exploration} proposes a Laplace approximation to the posterior, whereas \citet{dwaracherla2021langevin, ishfaq2023provable, bayesian2024vaart, azizzadenesheli2018efficient} use different MCMC samplers. All these algorithms build upon the typical DQN agent.

For this work we build upon Parallel Q-learning (PQN), which is a more modern and performant baseline Q-learning algorithm. We use Watkins' Q-estimator \citep{watkins1992q}, which is a small modification to PQN's $\text{Peng}(\lambda)$ to accommodate for the off-policy samples that we get from Thompson sampling. Our agent then uses 
$$\log p(\theta | \mathcal{D}) \propto \sum_{\ \ \mathclap{(s, a, r, s') \in \mathcal{D}}} \log p_{\mathit{TD}}(\mathit{TD}(\theta, s, a, r, s') | \theta) + \log p(\theta)$$ 
as target distribution, where $p_{\mathit{TD}}$ is the likelihood of a TD error and $p(\theta)$ is the chosen prior. The likelihood is estimated from minibatches of data. We initially assume the temporal difference errors are normally distributed with standard deviation $\sigma_{\mathit{TD}}$, which we treat as a hyperparameter.
    
As inference method, we use Gradient Guided Monte Carlo (GGMC) \citep{garriga2021exact}, which is a modern MCMC sampler from a family 
that are known to have good performance in supervised learning \citep{wenzel2020good}.  We modify the implementation to be compatible with Optax, and translate the hyperparameters to result in equivalent learning speed of Adam. Finally, to improve the mixing of our MCMC sampler, we run an ensemble of 10 chains in parallel, making the final architecture similar to ensemble-based Q-learning methods such as BootDQN \citep{osband2016deep, osband2018randomized}. In line with prior work, we remove $\epsilon$-greedy exploration from our agent and implement Thompson sampling by sampling one model at the start of a training batch, and acting greedily with this model for multiple steps before sampling a new one.

\subsection{Experimental Setup}

\paragraph{Improved Priors} Using the same hyperparameters for Bayesian DQN as our previous experiments in Section~\ref{sec:coldposterior-dqn}, we swap out the prior with both a Laplace distribution and a Learned prior. For the Laplace distribution, we rescale the scale parameter to match the standard deviation of a normal distribution. For the learned prior we use no rescaling. The learned prior is a separate normalizing flow for each neural network layer, and is trained to fit the empirical distribution displayed in Figure~\ref{fig:priors_all} by aggregating the parameters over all environments except those of MinAtar, which we leave as testing environments similar to the typical train-test split in supervised learning. We refer to Appendix~\ref{sec:normflows} for more details on the normalizing flow architecture.

\paragraph{Learned Likelihoods}
We fit a small normalizing flow to the temporal difference errors observed by a pre-trained PQN agent, creating a separate, environment-specific model for each MinAtar environment. These models serve as oracles that capture the empirical distribution of TD errors under a near-optimal policy. We then run our Bayesian DQN agent from scratch, but  replacing the Gaussian likelihood with the density of the model of the corresponding environment.

\begin{figure*}[t]
    \centering
    \includegraphics[width=0.9\linewidth]{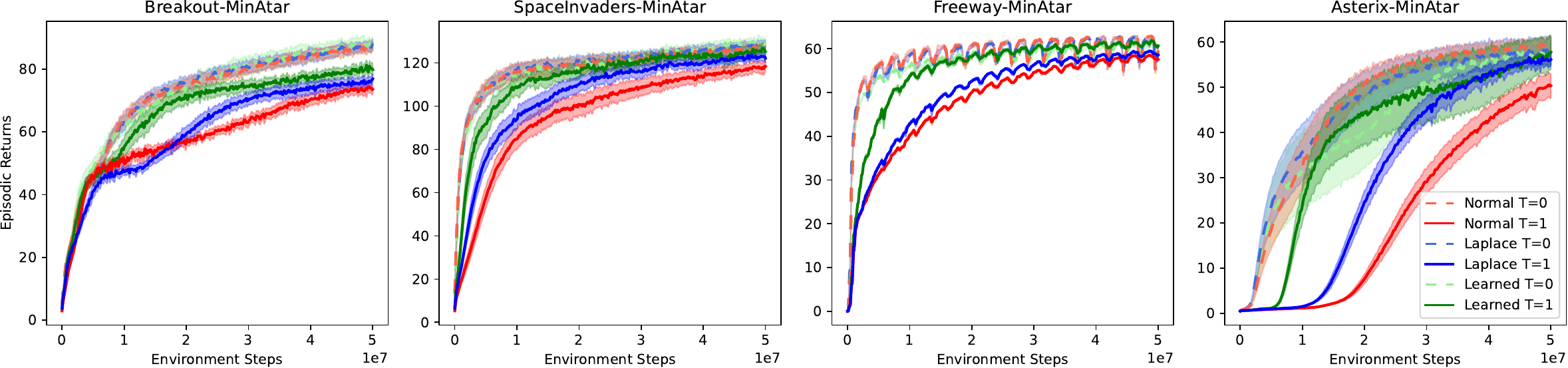}
    \caption{Cumulative returns of 
    Bayesian agent with a learned prior, Laplace prior and normal prior both for $T=1$ and $T=0$ (MAP). Lines are the mean of 30 seeds with shaded areas denoting one standard error.}
    \label{fig:learned-laplace-normal}
\end{figure*}

\subsection{Numerical Results}
\label{sec:results}
\paragraph{Improved Priors}
We can see in Figure~\ref{fig:learned-laplace-normal} that improving the prior distribution can significantly improve performance of a Bayesian DQN agent. Using a Laplace prior, which is only a tiny code difference and practically no extra computational cost is already significantly better than using a normal distribution, even at the hyperparameters for which the agent with the normal prior was tuned. 

Furthermore, Figure~\ref{fig:learned-laplace-normal} shows that the meta-learned prior improves performance once again, almost closing the cold posterior gap in SpaceInvaders, Freeway and Asterix. The fact that this prior was fit to the neural network parameters on Gymnax environments unrelated to MinAtar highlights that it is possible to develop priors that \emph{generalize} over environments. While this prior distribution is more involved from a programming standpoint, the computational burden is not significantly increased due to maintaining the i.i.d.\@ assumption with neural network layers. Interestingly, improving the prior appears to have little effect on the agent with $T=0$, indicating that the prior can aid in mitigating the cold posterior effect but does not provide better regularization in a maximum likelihood setting.

\begin{figure}[t]
    \centering
    \includegraphics[width=0.90\linewidth]{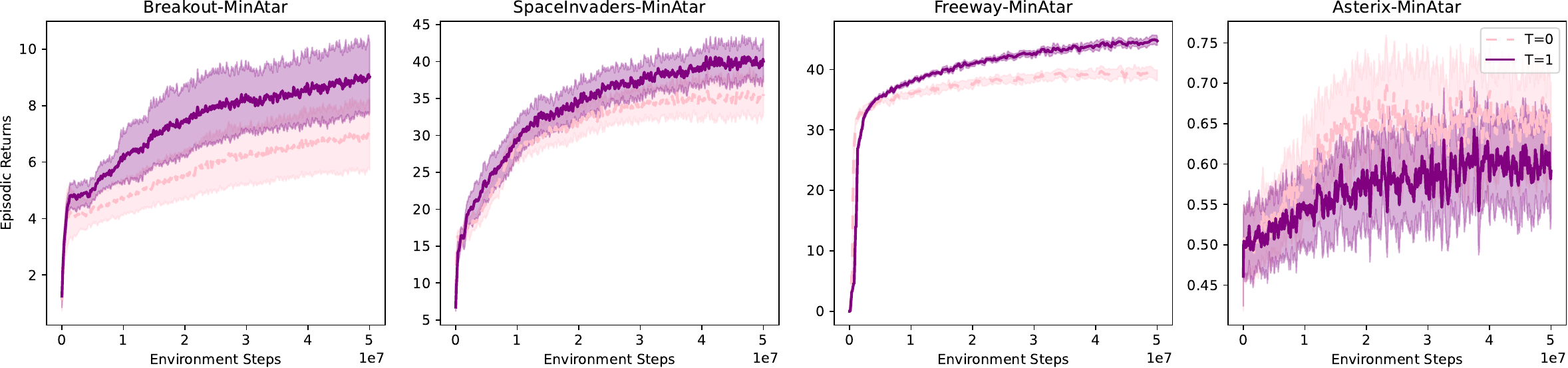}
    \caption{Return curves for Bayesian DQN at $T=0$ and $T=1$ with meta-learned prior and learned likelihoods on 
    MinAtar. Lines are the mean of 10 independent seeds, with shaded area denoting one standard error. 
    }
    \label{fig:minatar-learned-likelihood}
\end{figure}

\paragraph{Learned Likelihoods}
Figure~\ref{fig:minatar-learned-likelihood} shows the performance of Bayesian DQN with learned priors and learned likelihoods at $T=0$ and $T=1$. On Asterix the agent fails to learn anything, while on the other environments the agent with $T=1$ outperforms the agent with $T=0$. While the untempered posterior outperforms the MAP estimate in these experiments, it should be noted that all methods in this plot significantly underperform our agents where only the prior is learned. The poor results for $T=0$ signify that the log-density of the empirical distribution leads to a poorly conditioned optimization problem, as predicted. We leave the development of likelihoods that are both realistic and easy to optimize for future research.

\section{Conclusion}
\label{sec:conc}

In this work, we have demonstrated that there exists a cold posterior effect in Deep Q-learning.
We investigated possible causes by statistically testing common assumptions on the likelihoods, as well as observing empirical distribution over parameters of trained agents to evaluate choices of priors. We show empirically that making better prior choices improves performance of Bayesian DQN agents, while keeping the performance of maximum likelihood approaches constant. Finally, we demonstrate that choosing likelihoods close to the true distribution closes the cold posterior effect. Together, our theoretical investigation and experimental results signify that there is significant room for improvement in Bayesian deep model-free RL, and more focus should be put on the likelihood and prior assumptions. We show that a principled change to a Laplace prior with a single line of code already yields significant improvements without retuning hyperparameters, and that fitting a prior to previously trained weights improves further and generalizes to new environments. A promising future research direction is to develop likelihoods that impose a smooth optimization landscape while being more realistic than the commonly assumed Gaussian.

\begin{ack}
This work has received funding from the European Union's Horizon 2020 research and innovation programme, under grant agreements 964505 (E-pi) and 952215 (TAILOR). 
\end{ack}

\medskip

{
\small
\bibliographystyle{abbrvnat}
\bibliography{bibliography}
}



\newpage
\appendix
\section{Experimental Details} \label{sec:hyperparameters}
\paragraph{Architecture} For our Bayesian DQN agent, we used the same architecture as PQN on the Gymnax environments. We flatten the input, followed by two hidden layers of size 128 with relu activations and layernorm, and finally a linear layer with one unit for each action.

\paragraph{Hyperparameters} We optimized the learning rate $\ell \in [5 \cdot 10^{-5}, 10^{-3}]$, standard deviation of the Gaussian prior $\sigma_p \in [10^{-2}, 10^{10}]$ and standard deviation of the Gaussian likelihood $\sigma_\mathit{TD} \in [10^2, 10^{-2}]$ to work well on Breakout-MinAtar at $T=1$. The rest of the hyperparameters we keep unchanged from the default PQN hyperparameters. After 90 trials of Bayesian search we settled for the MinAtar hyperparameters displayed in Table~\ref{tab:hyperparameters}. For Deep Sea specifically we ran an extra grid search for the likelihood standard deviation and found $\sigma_l = 0.1$ to work well environment size 20.

For our regular PQN agent experiments to empirically investigate priors and TD error distributions, we use the default hyperparameters of PQN that were tuned for Gymnax.

\paragraph{Priors \& Likelihoods} When testing Laplace priors, we did not retune any hyperparameters, but instead rescaled the scale parameter $\sigma_p$ to match the standard deviation of the tuned Gaussian. For our experiments with learned priors, we simply plug in the learned prior without any rescaling. We also use unscaled learned priors for our experiments with learned likelihoods.

\paragraph{MCMC Sampler} For the GGMC damping $a$ and step size $h$ parameters, we translated the default Adam \citep{kingma2014adam} parameter $\beta_1 = 0.9$ and our standard learning rate by $a = \exp(-(1-\beta_1))$ and $h=\sqrt{(1-\beta_1)\frac{\ell}{n_\mathtt{data}}}$, where $n_\texttt{data}$ denotes the number of environment transitions the agent has observed. In contrast to \citet{garriga2021exact}, we fold $\sqrt{(1-\beta_1)}$ into the step size $h$ as this more closely matched the update sizes of Adam at the same parameters. We update $n_\mathtt{data}$ for every batch of collected trajectories, and rescale the the mean likelihood of a batch by $n_\mathtt{data}$ to reflect the full data set size.

\begin{table}[b]
    \centering
    \begin{tabular}{|c|c|c|}
    \hline
    Name & Symbol & Value\\
    \hline
     Learning rate & $\ell$ & $10^{-3}$ \\
     Prior scale & $\sigma_p$ & $1.679$ \\
     Likelihood scale & $\sigma_\mathit{TD}$ & $ 0.56$ (MinAtar), $0.1$ (Deep Sea) \\
     Ensemble size & - & $10$ \\
     GGMC damping& $a = \exp(-(1-\beta_1))$ & $\exp(-0.1)$\\
     GGMC step size & $h = \sqrt{(1 - \beta_1)\frac{\ell}{{n_\texttt{data}}}}$ & $\sqrt{\frac{10^{-4}}{n_\texttt{data}}}$\\
     GGMC Preconditioner Decay & $\beta_2$ & $0.999$\\
     \hline
    \end{tabular}
    \caption{Hyperparameters of our Bayesian DQN agent}
    \label{tab:hyperparameters}
\end{table}
\section{Normalizing Flow Details}\label{sec:normflows}
A normalizing flow is a parameterized invertable mapping $f_\psi: Z \rightarrow X$, that together with a base distribution $p_z(z)$, forms a pushforward distribution on $X$:
\begin{equation}\label{eq:norm-flow}
    p_\psi(x) = p_z(f^{-1}_\psi(x))|D_{f_{\psi}}(f^{-1}_\psi(x))|^{-1},
\end{equation}
where $|\ .\ |$ denotes the determinant and $D_{f_\psi}$ denotes the Jacobian of $f_\psi$. For more details, we refer to \citet{rezende2015variational}. For our work, it is most important that normalizing flows are a flexible variational inference framework, and that we can optimize the parameters $\psi$ so that $p_\psi(x)$ matches our desired distribution. Furthermore, the invertability of $f_\psi$ allows us to exactly evaluate the density $p_\psi(x)$ via Equation~\ref{eq:norm-flow}, which is crucial for our application as we want the flow to take the place of a prior or likelihood, which we need to evaluate to perform inference. Normalizing flows are typically constructed precisely so that evaluation of the log-density is cheap to compute by using operations that have simple Jacobians.

For all our normalizing flows, we used a single rational quadratic spline with two knots from the Distrax package, transforming a standard normal distribution to the target distribution. We also include affine rescaling before and after the spline. This means that each normalizing flow has only $7\ (\text{splines})+ 2\ (\text{affine}) + 2\  (\text{affine})= 11$ parameters, making them cheap to fit and evaluate, while being much more expressive than predefined distributions.

The normalizing flow $p_{\psi}(x)$ is trained using Adam \citep{kingma2014adam} to minimize the KL-divergence between the flow and the samples
$$ \arg \min_\psi KL(p_x, p_{\psi}) \propto \arg \min \mathbb{E}_x( -p_{\psi}(x)),$$
where $p_x$ denotes the empirical distribution and $x$ are the samples, which are neural network weights in the case of our prior experiments and TD errors in the case of our likelihood experiments.

In the experiments regarding priors, we fit an independent model to each neural network layer aggregated over all environments excluding MinAtar. Each weight in the layer shares the same normalizing flow, and weights are assumed to be drawn i.i.d. from this flow as a prior.

For the likelihood experiments, we fit the same normalizing flow architecture to the TD errors of each environment individually.

\section{Kolmogorov-Smirnov Test}\label{sec:ks-test}
The Kolmogorov-Smirnov (KS) test \citep{daniel1990applied} is a common statistical test to check whether two distributions are the same. The statistic for two cumulative density functions (cdf) $F_1$ and  $F_2$, is defined as $$D = \sup_x|F_1(x) - F_2(x)|,$$
which is the maximum deviation between the two cumulative densities. The statistic $D$ can then be compared to a critical value $D_p$ that depends on the significance level $p$ to decide whether the hypothesis should be rejected, i.e. the distributions are not the same.

We apply the KS test to an empirical sample of $N=8192$ TD errors collected by PQN \emph{after} training for 50 million samples, and compare to both a normal distribution and a Laplace distribution. To make our tests invariant to location and scale, we normalize the TD-errors $\epsilon_i$ by \begin{equation}\label{eq:test-norm}
    \hat \epsilon_i =  \frac{\epsilon_i - \bar \epsilon}{\sigma_{\epsilon}}
\end{equation} where $\bar \epsilon$ and $\sigma_\epsilon$ are the empirical mean and standard deviation. We then construct the empirical cdf $F_{\hat \epsilon}$ and compute the test statistic $D$ with respect to both the cdfs of a Gaussian and Laplace distribution with mean 0 and scale parameters 1 and $\frac{1}{\sqrt{2}}$ respectively, which corresponds to variances of $1$ for both distributions

To compute the critical values for both our test statistics, we simulate $D$ under the nulhypothesis 10000 times, each time by sampling $N=8192$ independent samples from a Gaussian and Laplace distribution, renormalizing them equivalently to Equation~\ref{eq:test-norm}, and storing the resulting KS statistics $D$. We then define the critical value for $p=0.05$ as the 0.05-th percentile of our simulation results. We repeated this entire experiment 10 times for each of the 19 environments to produce the left plot in Figure~\ref{fig:likelihoods_compare}.
\end{document}